\documentclass[letterpaper, 10 pt, journal, twoside]{ieeetran} 

\IEEEoverridecommandlockouts

\usepackage{multirow}
\usepackage{graphicx}
\usepackage{comment}
\usepackage{cite}
\usepackage{amsmath}
\usepackage{amsfonts}
\usepackage{mathtools}
\usepackage{tabularx}
\usepackage{graphicx}
\usepackage{array}
\usepackage{float}
\usepackage{algorithmic}
\usepackage[linesnumbered, ruled, vlined]{algorithm2e}
\usepackage[colorlinks,urlcolor=blue]{hyperref}
\usepackage{setspace}

\SetCommentSty{mycommfont}

\DeclareMathOperator*{\argmin}{arg\,min}

\title{\LARGE \bf Direct LiDAR Odometry: \\ 
Fast Localization with Dense Point Clouds}

\author{Kenny Chen$^{1}$, Brett T. Lopez$^{2}$, Ali-akbar Agha-mohammadi$^{3}$, and Ankur Mehta$^{1}$%
\thanks{Manuscript received: September 9, 2021; Revised December 1, 2021; Accepted December 23, 2021.}%
\thanks{This paper was recommended for publication by
Editor Sven Behnke upon evaluation of the Associate Editor and Reviewers' comments.
This work was partially conducted at NASA JPL in support of the DARPA Subterranean Challenge.}%
\thanks{$^{1}$Kenny Chen and Ankur Mehta are with the Department of Electrical and Computer Engineering, University of California Los Angeles, Los Angeles, CA, USA. {\tt\footnotesize \{kennyjchen, mehtank\}@ucla.edu}}%
\thanks{$^{2}$Brett T. Lopez is with the Department of Mechanical and Aerospace Engineering, University of California Los Angeles, Los Angeles, CA, USA. {\tt\footnotesize btlopez@ucla.edu}}%
\thanks{$^{3}$Ali-akbar Agha-mohammadi is with the NASA Jet Propulsion Laboratory, California  Institute  of  Technology, Pasadena, CA, USA. {\tt\footnotesize  aliagha@jpl.nasa.gov}}%
\thanks{Digital Object Identifier (DOI): see top of this page.}%
}

\begin{document}
\bstctlcite{IEEEexample:BSTcontrol}
\IEEEaftertitletext{\vspace{-4mm}}
\maketitle

\markboth{IEEE Robotics and Automation Letters. Preprint Version. Accepted January, 2022}
{Chen \MakeLowercase{\textit{et al.}}: Direct LiDAR Odometry: Fast Localization with Dense Point Clouds}


\begin{abstract}

Field robotics in perceptually-challenging environments require fast and accurate state estimation, but modern LiDAR sensors quickly overwhelm current odometry algorithms. To this end, this paper presents a lightweight frontend LiDAR odometry solution with consistent and accurate localization for computationally-limited robotic platforms. Our Direct LiDAR Odometry (DLO) method includes several key algorithmic innovations which prioritize computational efficiency and enables the use of dense, minimally-preprocessed point clouds to provide accurate pose estimates in real-time. This is achieved through a novel keyframing system which efficiently manages historical map information, in addition to a custom iterative closest point solver for fast point cloud registration with data structure recycling. Our method is more accurate with lower computational overhead than the current state-of-the-art and has been extensively evaluated in multiple perceptually-challenging environments on aerial and legged robots as part of NASA JPL Team CoSTAR's research and development efforts for the DARPA Subterranean Challenge.

\end{abstract}



\section{Introduction}

\IEEEPARstart{A}{ccurate} state estimation and mapping in large, perceptually-challenging environments have become critical capabilities for autonomous mobile robots. Whereas typical visual SLAM approaches often perform poorly in dust, fog, or low-light conditions, LiDAR-based methods can provide more reliable localization due to the superior range and accuracy of direct depth measurements \cite{cadena2016past}. However, recent work on LiDAR odometry (LO) have revealed the challenges of processing the large number of depth returns generated by commercial LiDAR sensors in real-time for high-rate state estimation \cite{shan2018lego, ebadi2020lamp}. This work presents several algorithmic innovations that make real-time localization with dense LiDAR scans feasible while also demonstrating the superiority of our method in terms of accuracy and computational complexity when compared to the state-of-the-art.

Current LO algorithms estimate a robot's egomotion in two stages: first, by performing a ``scan-to-scan" alignment between adjacent LiDAR frames to recover an immediate motion guess, followed by a ``scan-to-map" registration between the current scan and past environmental knowledge to increase global pose consistency. Unfortunately, the large number of data points per scan from modern LiDARs quickly overwhelms computationally-limited processors and bottlenecks performance during alignment, which can induce frame drops and ultimately cause poor pose estimation. More specifically, scan-to-scan alignment requires a registration of corresponding points between two clouds, but this process often involves a nearest-neighbor search which grows exponentially with the number of points per scan. Feature-based methods \cite{shan2018lego, lvisam2021shan, ye2019tightly, shan2020lio} attempt to mitigate this by using only the most salient points, but these methods employ an often computationally-intensive feature extraction step and may accidentally discard data which could otherwise help improve the quality of downstream registration. Moreover, in scan-to-map alignment, keyed environmental history (which consists of all or a subset of past points) grows rapidly in size as new scans are acquired and stored in memory. While aligning with a submap (rather than the full history of scans) helps increase computational efficiency, the perpetual addition of points still significantly expands the nearest-neighbor search space for typical submap extraction methods. Tree-based data structures have been shown to decrease this nearest-neighbor search cost significantly \cite{bhatia2010survey}, but the extraction of a local submap still involves too many points after just a few keyframes, thus preventing consistent performance for long-term navigation.

\begin{figure}[!t]
    \centering
    \includegraphics[width=0.95\columnwidth]{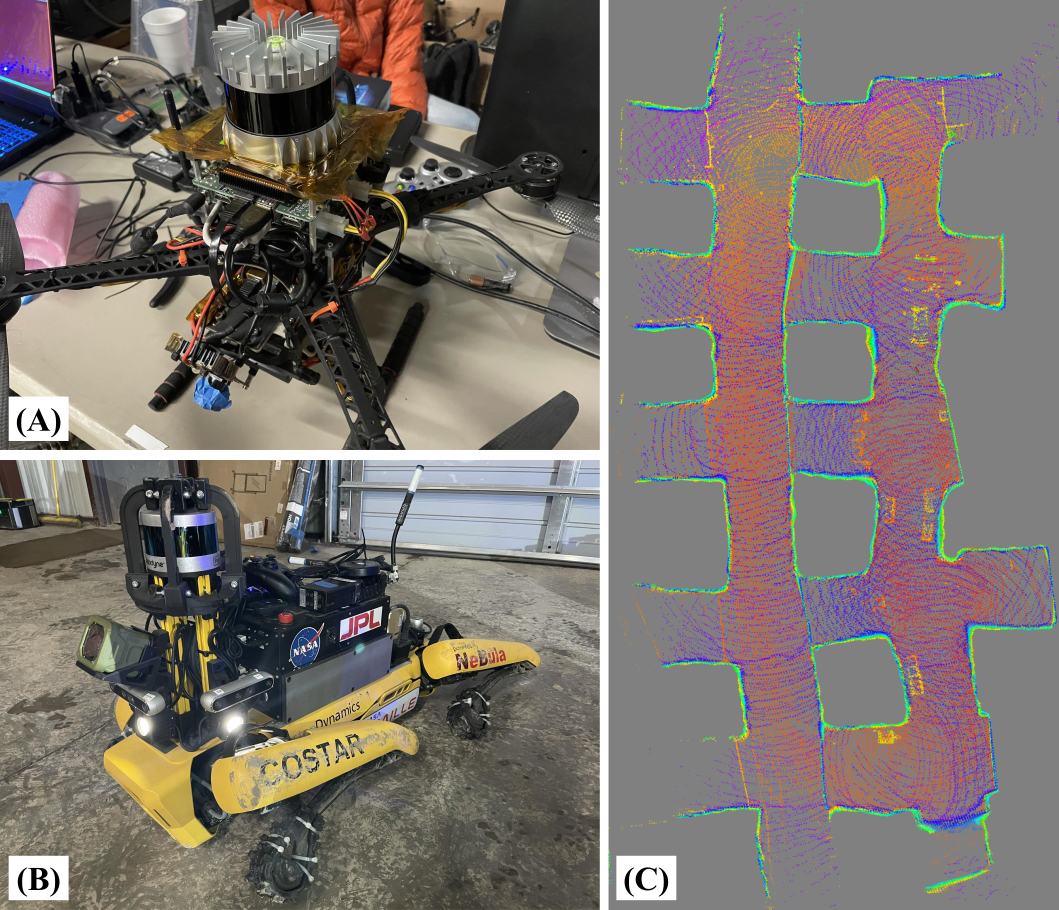}
    \vspace{-2mm}
    \caption{\textbf{Fast and lightweight LiDAR odometry.} Two of Team CoSTAR's robotic platforms which have limited computational resources. (A) Our custom quadrotor platform which features an Ouster OS1 LiDAR sensor on top. (B) A Boston Dynamics Spot robot with a mounted custom payload and a Velodyne VLP-16 with protective guards. (C) Top-down view of a mapped limestone mine using our lightweight odometry method on these robots during testing and integration for the DARPA Subterranean Challenge.}
    \label{fig:header}
    \vspace{-2mm}
\end{figure}

\begin{figure*}[!t]
    \centering
    \vspace{4mm}
    \includegraphics[width=0.8\textwidth]{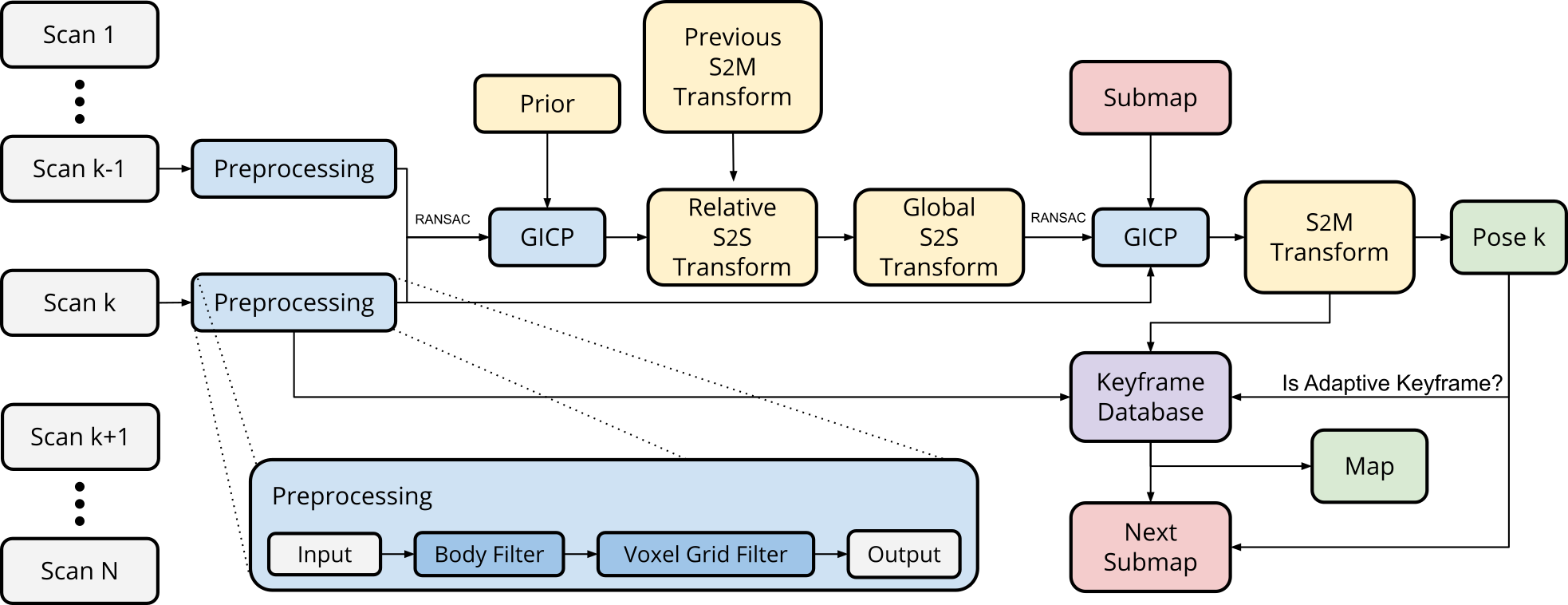}
    \vspace{-2mm}
    \caption{\textbf{LiDAR odometry architecture.} Our system first retrieves a relative transform between two temporally-adjacent scans of times $k$ and $k-1$ through scan-to-scan (S2S) matching with RANSAC outlier rejection and an optional rotational prior from IMU. This initial estimate is then propagated into the world frame and used as the initialization point for our secondary GICP module for scan-to-map optimization (S2M), which scan-matches the current point cloud $\mathcal{P}_k$ with a derived submap $\mathcal{S}_k$ consisting of scans from nearby and boundary keyframes. The output of this is a globally-consistent pose estimate which is subsequently checked against several metrics to determine if the current pose should be stored as a new keyframe.}
    \label{fig:architecture}
    \vspace{-6mm}
\end{figure*}

In this letter, we present our Direct LiDAR Odometry (DLO) algorithm, a high-speed and computationally-efficient frontend localization solution which permits the direct use of dense point cloud scans without significant preprocessing. The main contribution of this work is a custom speed-first pipeline which accurately resolves robot egomotion in real-time using minimally-preprocessed LiDAR scans and an optional IMU on consumer-grade processors. A key insight of our work is the link between algorithmic speed and accuracy, and our approach is comprised of three core innovations. First, an adaptive keyframing system which efficiently captures significant environmental information through a novel spaciousness metric. Second, a fast keyframe-based submapping approach via convex optimization which generates permissive local submaps for global pose refinement. Third, NanoGICP, a custom iterative closet point solver for lightweight point cloud scan-matching with data structure recycling to eliminate redundant calculations. Our method has been extensively evaluated in numerous challenging environments on computationally-limited robotic platforms as part of Team CoSTAR's research and development efforts for the DARPA Subterranean Challenge, and we have open-sourced our code for benefit of the community\footnote{\color{blue}\href{https://github.com/vectr-ucla/direct\_lidar\_odometry}{https://github.com/vectr-ucla/direct\_lidar\_odometry}}.


\subsection*{Related Work}
LiDAR-based odometry is typically cast as a nonlinear optimization problem to calculate a best-fit homogeneous transform that minimizes the error across corresponding, i.e., matching, points and/or planes between two point clouds. Since correspondences are not known \textit{a priori}, techniques such as the iterative closest point (ICP) algorithm \cite{chen1992object} or other variants like Generalized ICP (GICP) \cite{segal2009generalized} have become the standard to align two point clouds; however, searching over all data points can be computationally costly. \textit{Feature}-based methods attempt to extract and use only the most salient points before scan-matching to reduce computation. Such features are found either via manually tuned methods \cite{zhang2014loam} or learned networks \cite{yew20183dfeat} and might consist of planes \cite{ye2019tightly}, lines and edges \cite{shan2020lio, lvisam2021shan}, or ground points\cite{shan2018lego}. These works aim to translate insights gained from visual odometry (VO) techniques into the 3D domain. However, adding this step increases computational overhead and risks discarding data points which could help with better correspondence matching for odometry accuracy. Alternatively, \textit{direct} methods attempt to align dense point clouds but must heavily downsample to achieve computational tractability \cite{nelson, palieri2020locus}. More recently, a recursive filtering framework, e.g. Kalman filter, has been proposed \cite{fastlio1, fastlio2} to achieve real-time performance but at the potential expense of estimation accuracy.

A second stage immediately following scan alignment between adjacent clouds has been shown to reduce global drift by increasing pose estimation consistency with previous past scans \cite{palieri2020locus, ebadi2020lamp}. In the scan-to-map stage, a scan-to-scan transformation is further refined by aligning the current point cloud with an existing in-memory map; this submap is typically derived by retrieving nearby map points within some radius of the robot's current position. However, this search in ``point-space" can quickly explode in computational expense due to the sheer number of operations needed to retrieve the nearest neighbor data points. While there exists techniques to mitigate this such as only incrementally storing map data at keyed locations \cite{shan2020lio}, this search still involves thousands of calculations which can increase overall processor load and hence the potential to drop frames.

To address these issues, our DLO algorithm is built around a ``speed-first" philosophy to permit the use of minimally-preprocessed point clouds and provide accurate pose estimates even for robots with limited computational resources (Fig.~\ref{fig:architecture}). The key contribution of our work lies in how we efficiently derive our submap for global refinement in scan-to-map matching. That is, rather than extracting points within a local vicinity of a robot's current position as most works do, DLO instead searches in keyframe-space by associating a scan's set of points with its corresponding keyframe position. The submap is subsequently constructed by concatenating the clouds from a subset of historic keyframes derived from nearby keyframes and those which make up the convex hull; this provides the current scan with both nearby and distant points in the submap to anchor to. In addition, a custom GICP solver enables extensive reuse of data structures across multiple solver instantiations to eliminate redundant operations across the two-stage process. Our system also optionally accepts an initialization prior from an IMU in a loosely-coupled fashion to further improve accuracy during aggressive rotational motions. The reliability of our approach is demonstrated through extensive tests on several computationally-limited robotic platforms in multiple challenging environments. This work was part of Team CoSTAR's research and development efforts for the DARPA Subterranean Challenge in support of NASA Jet Propulsion Laboratory's Networked Belief-aware Perceptual Autonomy (NeBula) framework \cite{agha2021nebula}, in which DLO was the primary state estimation component for our fleet of autonomous aerial vehicles (Fig.~\ref{fig:header}A).


\section{Methods}

\subsection{Notation}
A point cloud, $\mathcal{P}$, is composed of a set of points $p \in \mathcal{P}$ with Cartesian coordinates $p_{i} \in \mathbb{R}^3$. We denote $\mathcal{L}$ as the LiDAR's coordinate system, $\mathcal{B}$ as the robot's coordinate system located at the IMU frame, and $\mathcal{W}$ as the world coordinate system which coincides with $\mathcal{B}$ at the initial position. Note that in this work we assume $\mathcal{L}$ and $\mathcal{B}$ reference frames coincide. Submap, covariance,  and kdtree structures are denoted as $\mathcal{S}$, $\mathcal{C}$ and $\mathcal{T}$, respectively. We adopt standard convention such that $x$ points forward, $y$ points left, and $z$ points upward, and our work attempts to address the following problem: given adjacent point clouds scans $\mathcal{P}_k$ and $\mathcal{P}_{k-1}$ at time $k$, estimate the robot's current pose $\hat{\textbf{X}}_k^{\mathcal{W}} \in \mathbb{SE}(3)$ and map $\mathcal{M}_k$ in $\mathcal{W}$.

\subsection{Preprocessing}
Our system assumes an input of 3D point cloud data gathered by a 360$^\circ$ LiDAR such as an Ouster OS1 (20Hz) or a Velodyne VLP-16 (10Hz). To minimize information loss from the raw sensor data, only two filters are used during preprocessing: first, we remove all point returns that may be from the robot itself through a box filter of size $1$m$^3$ around the origin. This is especially important if an aerial robot's propellers (Fig.~\ref{fig:header}A) or protective guards (Fig.~\ref{fig:header}B) are in the LiDAR's field of view. The resulting cloud is then sent through a 3D voxel grid filter with a resolution of $0.25$m to lightly downsample the data for subsequent tasks while maintaining dominate structures within the surrounding environment. Note that in this work we do not correct for motion distortion since non-rigid transformations can be computationally burdensome, and we directly use the dense point cloud rather than extracting features as most works do. On average, each cloud contains ${\sim}10{,}000$ points after preprocessing.

\vspace{1mm}
\begin{algorithm}[!t]
    \small
    \setstretch{0.8}

	\SetAlgoLined
	\textbf{input:} $\mathcal{P}_k$, $\hat{\textbf{X}}^{\mathcal{W}}_{k-1}$ ; \textbf{initialize:} $\hat{\textbf{X}}^{\mathcal{W}}_{k-1}$ = $\textbf{I}$ or gravityAlign() \\
	\textbf{output:} $\hat{\textbf{X}}^{\mathcal{W}}_{k}$, $\mathcal{M}_k$\\

	\While{$\mathcal{P}_k \neq \emptyset$} {

	    \small {\tcp{preprocessing}}
	    $\mathcal{\bar{P}}_k$ $\leftarrow$ preprocessPoints($\mathcal{P}_k$) ; \\
	    computeAdaptiveParameters($\mathcal{\bar{P}}_k$) ; \\
	    
	    \small {\tcp{initialization}}
	    \If {$k = 0$} {
	        $\mathcal{T}^{\text{t}_1}_k$, $\mathcal{C}^{\text{t}_1}_k$ $\leftarrow$ NanoGICP$_1$.build($\mathcal{\bar{P}}_k$) ; \\
	        $\mathcal{K}_k$ $\leftarrow$ updateKeyframeDatabase($\hat{\textbf{X}}^{\mathcal{W}}_{k-1}$, $\mathcal{\bar{P}}_k$) ; \\
	        \textbf{continue}; \\
	    }
	    
	    \small {\tcp{prior}}
	    \leIf {IMU} {
            $\tilde{\textbf{X}}^{\mathcal{L}}_k$ $\leftarrow$ $\tilde{\textbf{X}}^{\mathcal{B}}_k$;
        } {
            $\tilde{\textbf{X}}^{\mathcal{L}}_k$ $\leftarrow$ $\textbf{I}$
        }
        
	    \small {\tcp{scan-to-scan}}
    	$\mathcal{T}^{\text{s}_1}_k$, $\mathcal{C}^{\text{s}_1}_k$ $\leftarrow$ NanoGICP$_1$.build($\mathcal{\bar{P}}_k$) ; \\

        $\hat{\textbf{X}}^{\mathcal{L}}_k$ $\leftarrow$ NanoGICP$_1$.align($\mathcal{T}^{\text{s}_1}_k$, $\mathcal{T}^{\text{t}_1}_k$, $\mathcal{C}^{\text{s}_1}_k$, $\mathcal{C}^{\text{t}_1}_k$, $\tilde{\textbf{X}}^{\mathcal{L}}_k$) ; \\
        $\tilde{\textbf{X}}^{\mathcal{W}}_k$ $\leftarrow$ $\hat{\textbf{X}}^{\mathcal{W}}_{k-1}$ $\hat{\textbf{X}}^{\mathcal{L}}_k$ ; \\
        
        \small {\tcp{scan-to-map}}
        $\mathcal{Q}_k$ $\leftarrow$ getKeyframeNeighbors($\hat{\textbf{X}}^{\mathcal{W}}_{k-1}$, $\mathcal{K}_k$) ; \\
        $\mathcal{H}_k$ $\leftarrow$ getKeyframeHulls($\hat{\textbf{X}}^{\mathcal{W}}_{k-1}$, $\mathcal{K}_k$) ; \\
        $\mathcal{S}_k$ $\leftarrow$ $\mathcal{Q}_k \oplus \mathcal{H}_k$ ; \\
        
        \leIf{$\mathcal{S}_k \neq \mathcal{S}_{k-1}$} {
        $\mathcal{T}^{\text{t}_2}_k$ $\leftarrow$ NanoGICP$_2$.build($\mathcal{S}_k$) ;
        }{
        $\mathcal{T}^{\text{t}_2}_k$ $\leftarrow$ $\mathcal{T}^{\text{t}_2}_{k-1}$
        }

        $\mathcal{T}^{\text{s}_2}_k$ $\leftarrow$ $\mathcal{T}^{\text{s}_1}_k$; $\mathcal{C}^{\text{s}_2}_k$ $\leftarrow$ $\mathcal{C}^{\text{s}_1}_k$; $\mathcal{C}^{\text{t}_2}_k$ $\leftarrow$ $\sum_n^N \mathcal{C}^{\mathcal{S}}_{k,n}$ ; \\
        
        $\hat{\textbf{X}}^{\mathcal{W}}_k$ $\leftarrow$ NanoGICP$_2$.align($\mathcal{T}^{\text{s}_2}_k$, $\mathcal{T}^{\text{t}_2}_k$, $\mathcal{C}^{\text{s}_2}_k$, $\mathcal{C}^{\text{t}_2}_k$, $\tilde{\textbf{X}}^{\mathcal{W}}_k$) ; \\
        
        \small {\tcp{update keyframe database and map}}
        $\mathcal{K}_k$ $\leftarrow$ updateKeyframeDatabase($\hat{\textbf{X}}^{\mathcal{W}}_{k}$, $\mathcal{\bar{P}}_k$) ; \\
        $\mathcal{M}_k$ $\leftarrow$ $\mathcal{M}_{k-1}$ $\oplus$ $\left\{ \mathcal{K}_k \setminus \mathcal{K}_{k-1} \right\}$ ; \\

        \small {\tcp{propagate data structures}}
        $\mathcal{T}^{\text{t}_1}_k$ $\leftarrow$ $\mathcal{T}^{\text{s}_1}_k$; $\mathcal{C}^{\text{t}_1}_k$ $\leftarrow$ $\mathcal{C}^{\text{s}_1}_k$ ; \\
        
        \Return $\hat{\textbf{X}}^{\mathcal{W}}_{k}$, $\mathcal{M}_k$ \\
	}
	\caption{Direct LiDAR Odometry}
	\label{alg:dlo}
\end{algorithm}

\subsection{Scan Matching via Generalized-ICP}
LiDAR-based odometry can be viewed as the process of resolving a robot's egomotion by means of comparing successive point clouds and point clouds in-memory to recover an $\mathbb{SE}(3)$ transformation, which translates to the robot's 6-DOF motion between consecutive LiDAR acquisitions. This process is typically performed in two stages, first to provide a best instantaneous guess, which is subsequently refined to be more globally consistent with previous keyframe locations.

\subsubsection{Scan-to-Scan}
In the first stage, the scan-to-scan matching objective is to compute a relative transform $\hat{\textbf{X}}^{\mathcal{L}}_k$ between a source $\mathcal{P}_{k}^{\text{s}}$ and a target $\mathcal{P}_{k}^{\text{t}}$ (where $\mathcal{P}_{k}^{\text{t}}$ = $\mathcal{P}_{k-1}^{\text{s}}$) captured in $\mathcal{L}$ where
\begin{equation}
    \hat{\textbf{X}}^{\mathcal{L}}_k = \argmin_{\textbf{X}^{\mathcal{L}}_k} \, \mathcal{E} \left( \textbf{X}^{\mathcal{L}}_k \mathcal{P}_{k}^{\text{s}}, \mathcal{P}_{k}^{\text{t}} \right) \,. \\
    \label{eq:s2s_1}
\end{equation}
\noindent The residual error $\mathcal{E}$ from GICP is defined as
\begin{equation}
    \mathcal{E} \left( \textbf{X}^{\mathcal{L}}_k \mathcal{P}_{k}^{\text{s}}, \mathcal{P}_{k}^{\text{t}} \right) = \sum_i^N d_i^\top \left( \mathcal{C}_{k,i}^{\text{t}} \, + \, \textbf{X}^{\mathcal{L}}_k \mathcal{C}_{k,i}^{\text{s}} \textbf{X}^{\mathcal{L}^\top}_k \right)^{-1} d_i \,, \\
    \label{eq:s2s_2}
\end{equation}
\noindent such that the overall objective for this stage is
\begin{equation}
    \hat{\textbf{X}}^{\mathcal{L}}_k = \argmin_{\textbf{X}^{\mathcal{L}}_k} \, \sum_i^N d_i^\top \left( \mathcal{C}_{k,i}^{\text{t}} \, + \, \textbf{X}^{\mathcal{L}}_k \mathcal{C}_{k,i}^{\text{s}} \textbf{X}^{\mathcal{L}^\top}_k \right)^{-1} d_i \,, \\
    \label{eq:s2s_3}
\end{equation}
\noindent for $N$ number of corresponding points between point clouds $\mathcal{P}_{k}^{\text{s}}$ and $\mathcal{P}_{k}^{\text{t}}$, where $d_i = p_i^{\text{t}} - \textbf{X}^{\mathcal{L}}_k p_i^{\text{s}}$, $p_i^{\text{s}} \in \mathcal{P}_{k}^{\text{s}}, p_i^{\text{t}} \in \mathcal{P}_{k}^{\text{t}}, \forall i$, and $\mathcal{C}_{k,i}^{\text{s}}$ and $\mathcal{C}_{k,i}^{\text{t}}$ are the corresponding estimated covariance matrices associated with each point $i$ of the source or target cloud, respectively. As will be further discussed in Section~\ref{sec:optimization_priors}, we can initialize the above objective function with a prior supplied by external sensors in an attempt to push the convergence towards a global minimum. That is, for Eq.~(\ref{eq:s2s_3}), if a prior $\tilde{\textbf{X}}^{\mathcal{B}}_k$ is available by means of IMU preintegration, we can set the initial guess $\tilde{\textbf{X}}^{\mathcal{L}}_k = \tilde{\textbf{X}}^{\mathcal{B}}_k$ to create a loosely-coupled system. If a prior is not available however, the system reverts to pure LiDAR odometry in which $\tilde{\textbf{X}}^{\mathcal{L}}_k$ = $\textbf{I}$ and relies solely on point cloud correspondence matching for this step.

\subsubsection{Scan-to-Map}
After recovering an initial robot motion estimate, a secondary stage of scan-to-map matching is performed and follows a similar procedure to that of scan-to-scan. However, rather than computing a relative transform between two instantaneous point clouds, the objective here is to further refine the motion estimate from the previous step to be more globally-consistent by means of matching with a local submap. In other words, the task here is to compute an optimal transform $\hat{\textbf{X}}_k^{\mathcal{W}}$ between the current source cloud $\mathcal{P}_k^\text{s}$ and some derived submap $\mathcal{S}_k$ such that
\begin{equation}
    \hat{\textbf{X}}^{\mathcal{W}}_k = \argmin_{\textbf{X}^{\mathcal{W}}_k} \, \mathcal{E} \left( \textbf{X}^{\mathcal{W}}_k \mathcal{P}_{k}^{\text{s}}, \mathcal{S}_{k} \right) \,. \\
    \label{eq:s2m_1}
\end{equation}
\noindent After similarly defining the residual error $\mathcal{E}$ from GICP as in Eq.~(\ref{eq:s2s_2}), the overall objective function for scan-to-map is
\begin{equation}
    \hat{\textbf{X}}^{\mathcal{W}}_k = \argmin_{\textbf{X}^{\mathcal{W}}_k} \, \sum_j^M d_j^\top \left( \mathcal{C}_{k,j}^{\mathcal{S}} \, + \, \textbf{X}^{\mathcal{W}}_k \mathcal{C}_{k,j}^{\text{s}} \textbf{X}^{\mathcal{W}^\top}_k \right)^{-1} d_j \,, \\
    \label{eq:s2m_2}
\end{equation}
\noindent for $M$ number of corresponding points between point cloud $\mathcal{P}_{k}^{\text{s}}$ and submap $\mathcal{S}_{k}$, where $\mathcal{C}_{k,j}^{\mathcal{S}}$ is the corresponding scan-stitched covariance matrix for point $j$ in the submap as defined later in Section~\ref{sec:algorithmic_impl}. Eq.~(\ref{eq:s2m_2}) is initialized using the propagated result from scan-to-scan in the previous section from $\mathcal{L}$ to $\mathcal{W}$, i.e.  $\tilde{\textbf{X}}^{\mathcal{W}}_k$ = $\hat{\textbf{X}}^{\mathcal{W}}_{k-1}$ $\hat{\textbf{X}}^{\mathcal{L}}_k$, so that this prior motion can be compared against historical map data for global consistency. The output of this stage $\hat{\textbf{X}}^{\mathcal{W}}_k$ is the final estimated robot pose used for downstream modules.

We note here that a key innovation of this work is how we derive and manage our submap for this stage. Whereas previous works create a submap by querying the locality of each individual point in a stored map, we associate scans to keyframes and search rather in keyframe-space to stitch point clouds together and create $\mathcal{S}_{k}$. The implications of this include a far faster and more consistent generation of a local submap, which is additionally more permissive as compared to a radius-based search and will be further discussed in Section~\ref{sec:submapping}.

\subsection{Optimization Prior}
\label{sec:optimization_priors}
Eq.~(\ref{eq:s2s_3}) describes the scan-to-scan nonlinear optimization problem and can be initialized with a prior to reduce the chances of converging into a sub-optimal local minima. This prior represents an initial guess of the relative motion between two LiDAR frames and can come from integrating angular velocity measurements from an inertial measurement unit (IMU). More specifically, angular velocity measurements $\hat{\boldsymbol{\omega}}_k$ is defined as $\hat{\boldsymbol{\omega}}_k = \boldsymbol{\omega}_k + \textbf{b}_k^{\boldsymbol{\omega}} + \textbf{n}_k^{\boldsymbol{\omega}}$ measured in $\mathcal{B}$ with static bias $\textbf{b}_k^{\boldsymbol{\omega}}$ and zero white noise $\textbf{n}_k^{\boldsymbol{\omega}}$ for convenience. After calibrating for the bias, a relative rotational motion of the robot's body between two LiDAR frames can be computed via gyroscopic propagation of the quaternion kinematics $\textbf{q}_{k+1} = \textbf{q}_k + (\tfrac{1}{2} \textbf{q}_k \otimes \boldsymbol{\omega}_k )\Delta t$. Here, $\textbf{q}_k$ is initialized to identity prior to integration, $\Delta t$ is the difference in time between IMU measurements in seconds, and only gyroscopic measurements found between the current LiDAR scan and the previous one are used. Note that we are only concerned with a rotational prior during IMU preintegration and leave the retrieval of a translational prior via the accelerometer for future work. The resulting quaternion of this propagation is converted to an $\mathbb{SE}(3)$ matrix with zero translational component to be used as $\tilde{\textbf{X}}^{\mathcal{B}}_k$, the scan-to-scan prior.

\subsection{Fast Keyframe-Based Submapping}
\label{sec:submapping}
\begin{figure}[!t]
    \centering
    \vspace{2mm}
    \includegraphics[width=0.85\columnwidth]{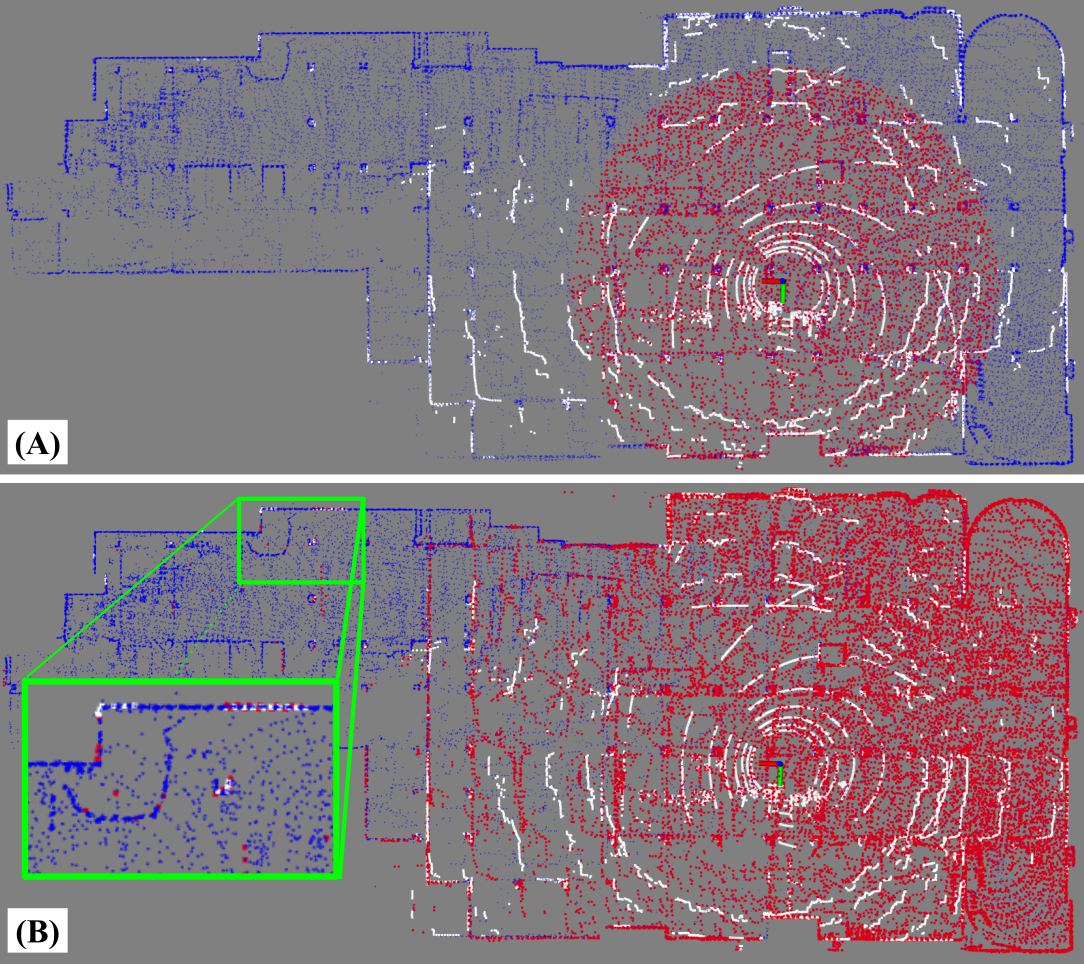}
    \vspace{-2mm}
    \caption{\textbf{Keyframe-based submapping.} A comparison between the different submapping approaches, visualizing the current scan (white), the derived submap (red), and the full map (blue). (A) A common radius-based submapping approach of $r$ = $20$m retrieved in point cloud-space. (B) Our keyframe-based submapping approach, which concatenates a subset of keyed scans and helps anchor even the most distant points in the current scan (green box) during the scan-to-map stage.}
    \label{fig:keyframe_comparison}
    \vspace{-2mm}
\end{figure}

A key innovation of this work lies in how our system manages map information and derives the local submap in scan-to-submap matching for global egomotion refinement. Rather than working directly with point clouds and storing points into a typical octree data structure, we instead keep a history of \textit{keyframes} to search within, in which each keyframe is linked to its corresponding point cloud scan in a key-value pair. The resulting local submap used for scan-to-submap matching is then generated by concatenating the corresponding point clouds from a subset of the keyframes, rather than directly retrieving local points within some radius of the robot's current position.

The implication of this design choice is twofold: first, by searching in ``keyframe-space" rather than ``point cloud-space," a much more computationally tractable problem is obtained. Radius-based searches within a cumulative point cloud map can require distance calculations against hundreds of thousands of points --- a process that quickly becomes infeasible even with an incremental octree data structure. Searching against keyframes, however, typically involves only a few hundred points even after long traversals and provides much more consistent computational performance, reducing the chances of dropping frames.
Additionally, a keyframe-based approach constructs a much more permissive submap as compared to range-based methods. That is, since the size of a submap derived from keyframe point clouds relies solely on the LiDAR sensor's range rather than a predetermined distance, the derived submap can have a larger overlap with the current scan; this is illustrated in Fig.~\ref{fig:keyframe_comparison}. In this example, a submap of fixed radius $r$ = $20$m insufficiently overlaps with the current scan and can introduce drift over time due to containing only spatially-nearby points; however, a keyframe-based approach covers most of the current scan which helps with better scan-to-map alignment. Expanding the radius size may help increase this overlap for radius-based methods, but doing so would significantly slowdown subsequent tasks such as the GICP covariance calculations.

\subsubsection{Keyframe Selection via kNN and Convex Hull}

\begin{figure}[!t]
    \centering
    \vspace{2mm}
    \includegraphics[width=0.85\columnwidth]{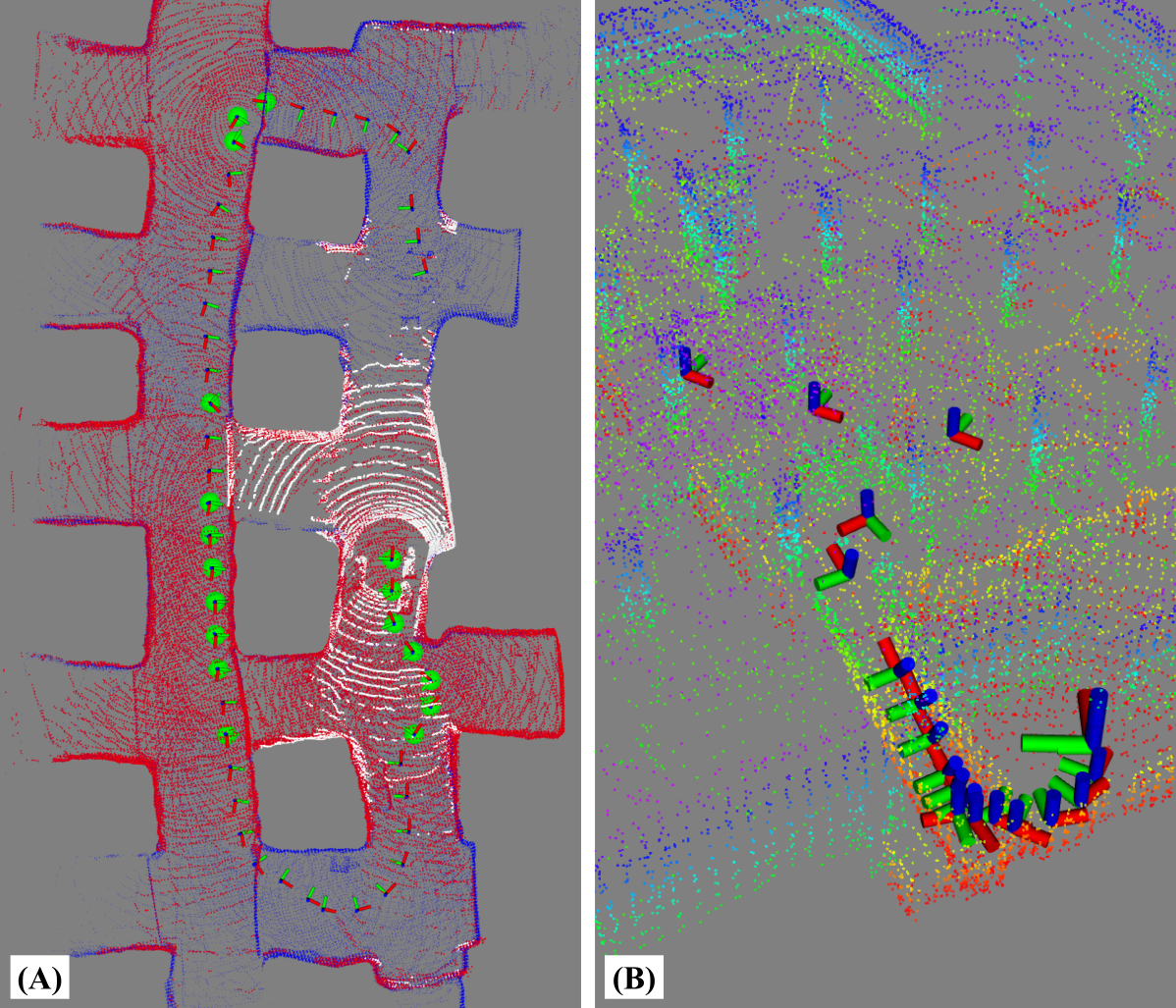}
    \vspace{-2mm}
    \caption{\textbf{Keyframe selection and adaptive thresholds.} (A) Our method's submap (red) is generated by concatenating the scans from a subset of keyframes (green spheres), which consists of $K$ nearest neighbor keyframes and those that construct the convex hull of the keyframe set. (B) An illustration of adaptive keyframing. In this scenario, the threshold decreases when traversing down a narrow ramp to better capture small-scale details.}
    \label{fig:convex_adaptive}
    \vspace{-2mm}
\end{figure}

To construct the submap $\mathcal{S}_k$, we concatenate the corresponding point clouds from a selected subset of environmental keyframes. Let $\mathcal{K}_k$ be the set of all keyframe point clouds such that $\mathcal{S}_k \subseteq \mathcal{K}_k$. We define submap $\mathcal{S}_k$ as the concatenation of $K$ nearest neighbor keyframe scans $\mathcal{Q}_k$ and $L$ nearest neighbor convex hull scans $\mathcal{H}_k$ such that $\mathcal{S}_k = \mathcal{Q}_k \oplus \mathcal{H}_k$, where the indices which specify the convex hull are defined by the set of keyframes which make up the intersection of all convex sets containing the keyframes which compose $\mathcal{K}_k$.

The result of this is illustrated in Fig.~\ref{fig:convex_adaptive}A, in which the keyframes highlighted in green are those that compose the extracted submap, indicated in red. Intuitively, extracting nearest neighbor keyframes aims to help with overlap of nearby points in the current scan, while those from the convex hull --- which contain boundary map points --- increase the overlap with more distant points in the scan. This combination reduces overall trajectory drift by maximizing scan-to-map overlap and provides the system with multiple scales of environmental features to align with. Note that keyframes which are classified as both a nearest neighbor and a convex hull index are only used once in the submap.

\subsubsection{Adaptive Keyframing}

The location of keyframes affects the derived submap and can subsequently influence accuracy and robustness of the odometry. Keyframe nodes are commonly dropped using fixed thresholds (e.g., every $1$m or $10^\circ$ of translational or rotational change) \cite{palieri2020locus, shan2020lio, lvisam2021shan}, but the optimal position can be highly dependent on a surrounding environment's structure. More specifically, in large-scale settings, features captured by the point cloud scan are much more prominent and can be depended on for longer periods of time. Conversely, for narrow or small-scale environments, a smaller threshold is necessary to continually capture the small-scale features (i.e., tight corners) in the submap for better localization. Thus, we choose to scale the translational threshold for new keyframes according to the ``spaciousness" in the instantaneous point cloud scan, defined as $m_k = \alpha m_{k-1} + \beta M_k$, where $M_k$ is the median Euclidean point distance from the origin to each point in the preprocessed point cloud, $\alpha$ = $0.95$, $\beta$ = $0.05$, and $m_k$ is the smoothed signal used to scale the keyframe threshold $th_k$ at time $k$ such that 
\begin{equation}
    th_k =
        \begin{cases}        
            10\text{m}   & \text{if } m_k > 20\text{m} \\
            5\text{m}    & \text{if } m_k > 10\text{m} \, \& \, m_k \leq 20\text{m} \\
            1\text{m}    & \text{if } m_k > 5\text{m} \, \& \, m_k \leq 10\text{m} \\
            0.5\text{m}  & \text{if } m_k \leq 5\text{m} \\
        \end{cases}
    \label{eq:thresh}
\end{equation}
with rotational threshold held fixed at $30^\circ$. Fig.~\ref{fig:convex_adaptive}B illustrates the effects of this adaptive thresholding, which helps with robustness to changing environmental dimension.

\subsection{Algorithmic Implementation}
\label{sec:algorithmic_impl}

\begin{table}[!t]
    \renewcommand{\arraystretch}{1.2}
    \vspace{2mm}
    \caption{Summary of Data Structure Recycling}
    \vspace{-5mm}
    \label{table:structs}
    \begin{center}
    \begin{tabular}{c||cc}
    Element & Scan-to-Scan & Scan-to-Map \\
    \hline
    $\mathcal{T}_k^{\text{source}}$ & \small build & $\xrightarrow{\text{reuse from S2S}}$ \\
    $\mathcal{T}_k^{\text{target}}$ & $\mathcal{T}_{k-1}^{\text{source}}$  & \small build when $\mathcal{S}_{k} \neq \mathcal{S}_{k-1}$  \\
    $\mathcal{C}_k^{\text{source}}$ & \small compute & $\xrightarrow{\text{reuse from S2S}}$ \\
    $\mathcal{C}_k^{\text{target}}$ & $\mathcal{C}_{k-1}^{\text{source}}$ & $\sum_n^N \mathcal{C}^{\mathcal{S}}_{k,n}$ \\
    \end{tabular}
    \end{center}
    \vspace{-2mm}
\end{table}

\subsubsection{Scan-Stitched Submap Normals}
Generalized-ICP involves minimizing the plane-to-plane distance between two clouds, in which these planes are modeled by a computed covariance for each point in the scan. Rather than computing the normals for each point in the submap on every iteration (which can be infeasible for real-time operation), we assume that the set of submap covariances $\mathcal{C}^{\mathcal{S}}_k$ can be approximated by concatenating the normals $\mathcal{C}^{\mathcal{S}}_{k,n}$ from $N$ keyframes which populate the submap such that $\mathcal{C}^{\mathcal{S}}_k \approx \sum_n^N \mathcal{C}^{\mathcal{S}}_{k,n}$. As a consequence, each submap's set of normals need not be explicitly computed, but rather just reconstructed by stitching together those calculated previously.

\subsubsection{Data Structure Recycling}
Expanding on the above, several algorithmic steps in current LiDAR odometry pipelines can benefit from data structure sharing and reuse, drastically reducing overall system overhead by removing unnecessary and redundant operations. As summarized in Table~\ref{table:structs}, the system requires eight total elements to successfully perform scan-to-scan and scan-to-map matching. This includes kdtrees $\mathcal{T}_k$ used to search for point correspondences and covariance matrices $\mathcal{C}_k$ for GICP alignment for both source and target clouds in each scan-matching process. 

Out of the four required kdtrees data structures, only two need to be built explicitly. That is, the tree for the source (input) cloud $\mathcal{T}_k^{\text{source}}$ can be built just once per scan acquisition and shared between both modules (as the same scan is used for both sources). For the scan-to-scan target tree $\mathcal{T}_k^{\text{target}}$, this is simply just the previous iteration's scan-to-scan source tree $\mathcal{T}_{k-1}^{\text{source}}$ and thus can be propagated. The scan-to-map target tree needs to be built explicitly, but since the submap is derived from a set of keyframes, this build only needs to be performed when the set of selected keyframes via our kNN and convex hull strategy changes from one iteration to the next, such that $\mathcal{S}_{k} \neq \mathcal{S}_{k-1}$. Otherwise, the data structure can just be reused again for additional computational savings. Point covariances $\mathcal{C}_k$ needed for GICP, on the other hand, only need to be computed once per scan aquisition, and its data can be shared directly in the other three instances.

\begin{figure}[!t]
    \centering
    \vspace{2mm}
    \includegraphics[width=0.8\columnwidth]{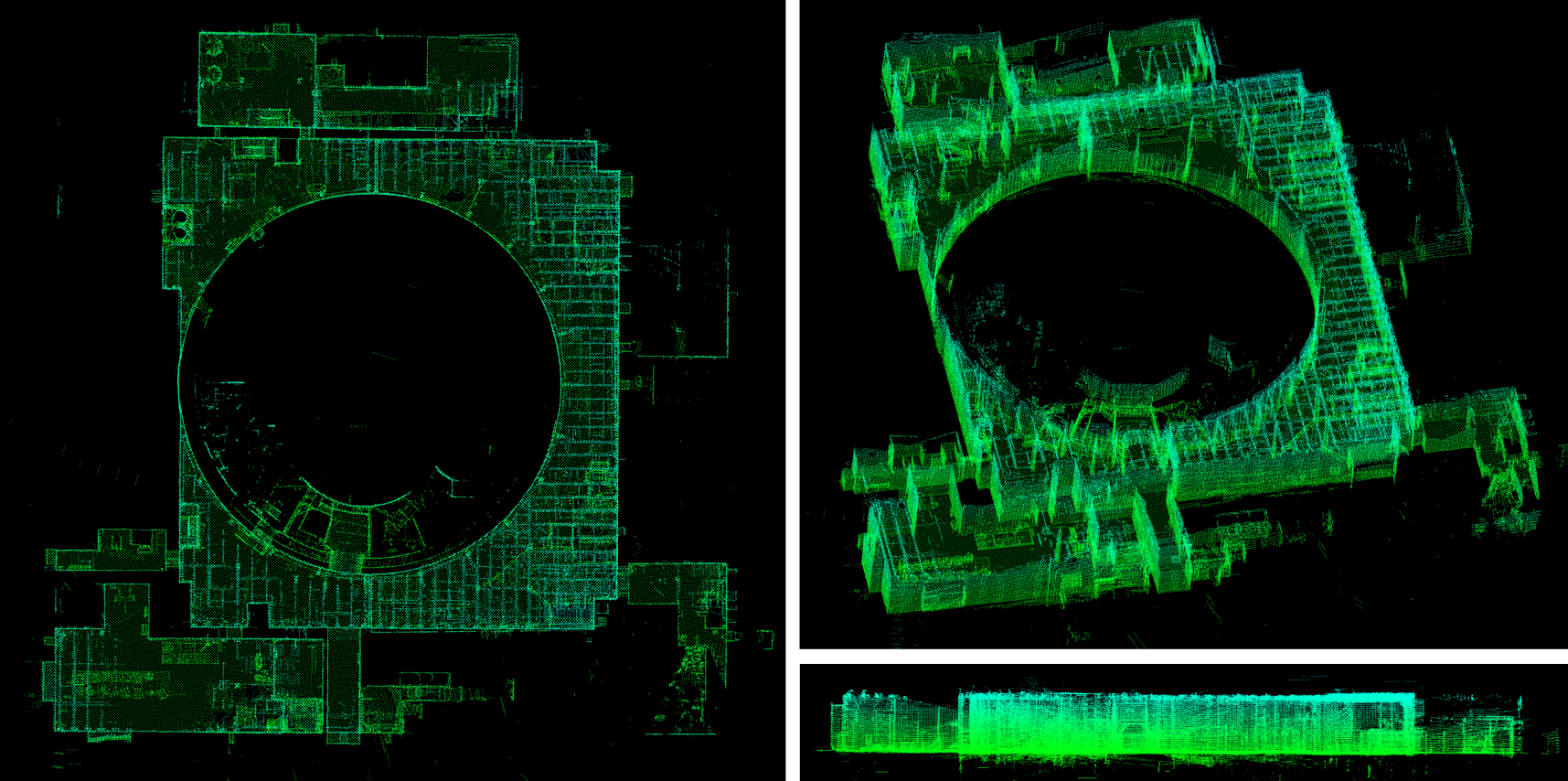}
    \vspace{-2mm}
    \caption{\textbf{Alpha course map.} Different views and angles of the dense 3D point cloud map generated using our DLO algorithm on the Urban Alpha dataset. Estimated positions at each timestamp were used to transform the provided scan into a world frame; this was performed for all scans across the dataset and concatenated / voxel filtered to generated the above images.}
    \label{fig:alpha}
    \vspace{-2mm}
\end{figure}

\begin{figure}[!t]
    \centering
    \includegraphics[width=0.85\columnwidth]{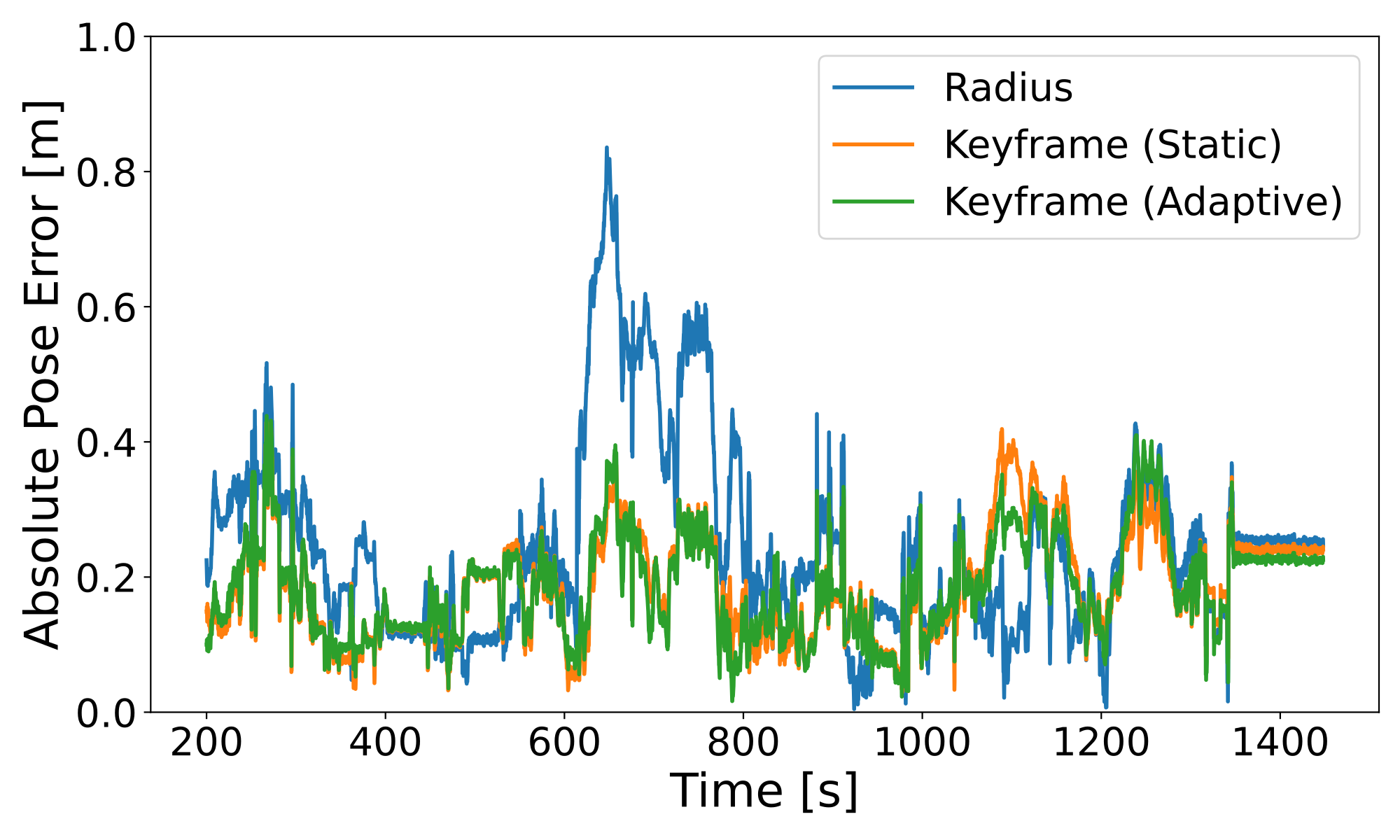}
    \vspace{-2mm}
    \caption{\textbf{Error comparison.} The absolute pose error plotted across a $1200$s window of movement, showing the difference between radius and keyframe submapping schemes. Keyframe-based approaches do not have the range restriction that radius-based approaches inherently contain, which directly translates to a lower error in odometry due to more perceptive submapping. Note that adaptive keyframing primarily helps with reliability and robustness to changes in environmental dimension (Fig.~\ref{fig:extreme}).}
    \label{fig:ape}
    \vspace{-2mm}
\end{figure}

\subsubsection{Dual NanoGICP}
To facilitate the cross-talking between scan-matching modules, we developed NanoGICP, a custom iterative closest point solver which combines the FastGICP\cite{koide2021voxelized} and NanoFLANN\cite{blanco2014nanoflann} open-source packages with additional modifications for data structure sharing as described before. In particular, NanoGICP uses NanoFLANN to efficiently build kdtree data structures, which are subsequently used for point cloud correspondence matching by FastGICP. In practice, data structure sharing is performed between two separate NanoGICP instantiations with different hyperparameters --- one to target each scan-matching problem --- and done procedurally as detailed in Algorithm~\ref{alg:dlo}.


\section{Results}

\begin{figure}[!t]
    \centering
    \vspace{2mm}
    \includegraphics[width=0.75\columnwidth]{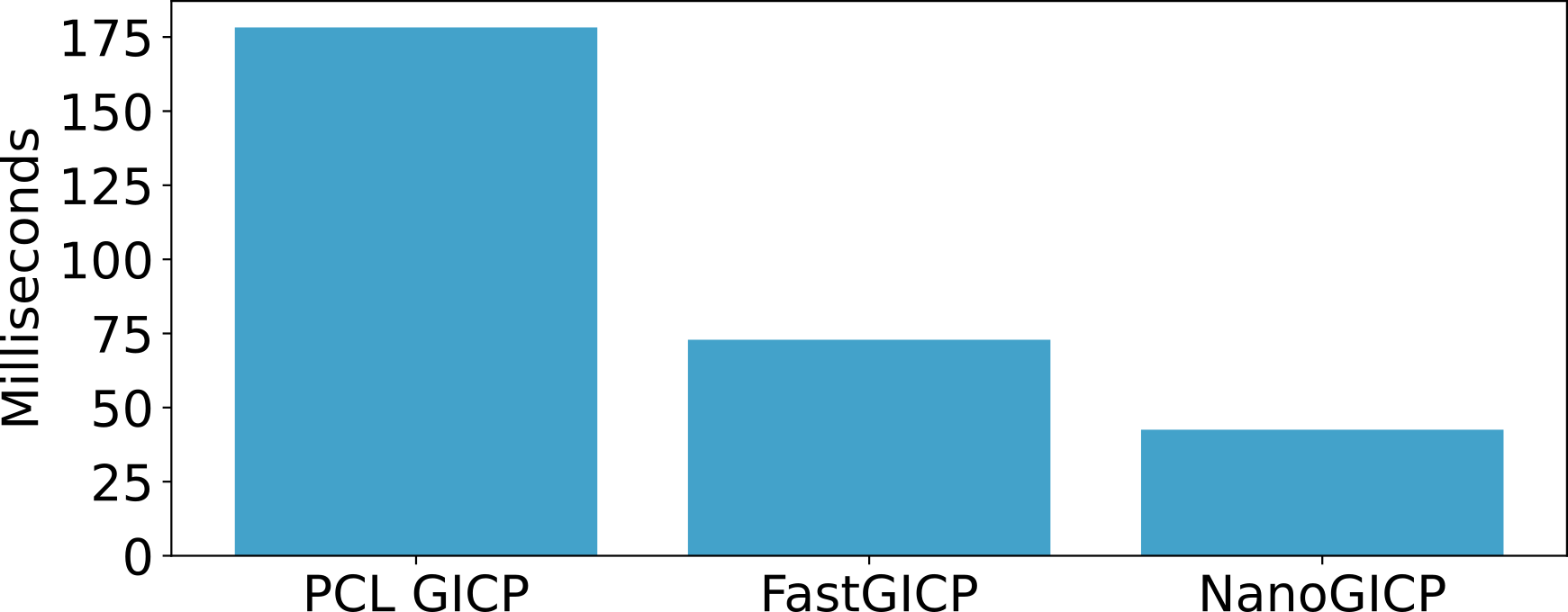}
    \vspace{-2mm}
    \caption{\textbf{Average convergence time.} A comparison of average convergence times across $100$ benchmark alignments for each algorithm, including our NanoGICP solver and two other open-source GICP packages.}
    \label{fig:gicp}
    \vspace{-2mm}
\end{figure}

\begin{figure}[!t]
    \centering
    \includegraphics[width=0.8\columnwidth]{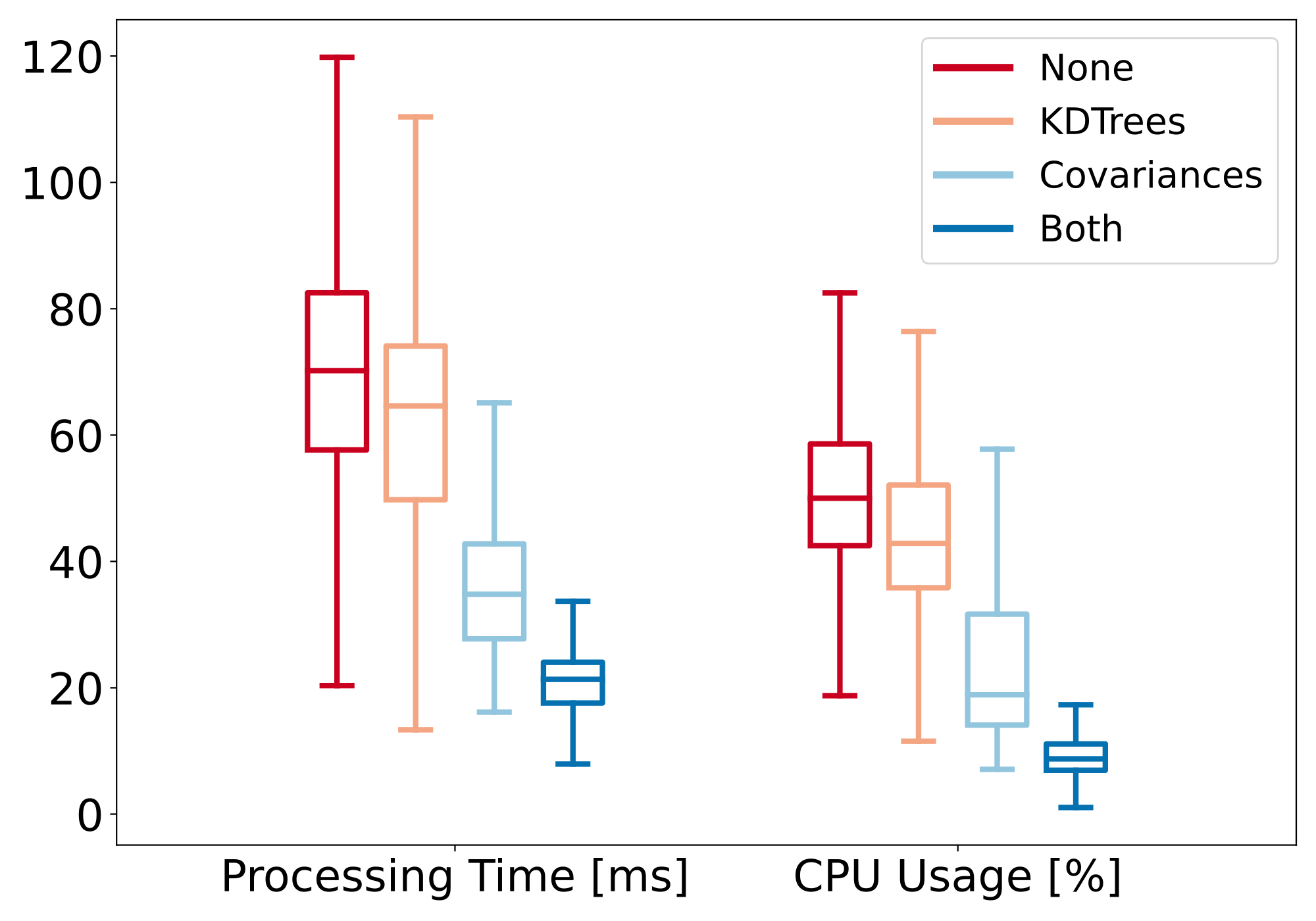}
    \vspace{-2mm}
    \caption{\textbf{Ablation study of data recycling schemes.} Box plots of the processing time and CPU usage for four different data recycling schemes, ranging from no data structure reuse to partial reuse and full reuse.}
    \label{fig:data_recycling}
    \vspace{-2mm}
\end{figure}

\begin{table}[!t]
    \renewcommand{\arraystretch}{1.5}
    \small
    \begin{center}
    \caption{Dropped LiDAR Scans per Recycling Scheme}
    \label{table:dropped_scans}
    \begin{tabular}{c||cccc}
     & None & KDTrees & Covariances & Both \\ \hline
    \% Scans & $9.37\%$ & $4.51\%$ & $0.00\%$ & $0.00\%$
    \end{tabular}
    \end{center}
    \vspace{-2mm}
\end{table}

\begin{table*}[!t]
    \vspace{2mm}
    \caption{Comparison on Benchmark Datasets}
    \vspace{-2mm}
    \label{table:comparison}
    \centering
    \setlength{\tabcolsep}{8 pt}
    \renewcommand{\arraystretch}{1.}
    \begin{tabular}{|l|c|c|c|c|c|c|c|c|c|c|}
    \hline
    \multicolumn{1}{|c|}{\multirow{3}{*}{Method}} & \multicolumn{4}{c|}{Alpha Course ($757.4$m)} & \multicolumn{4}{c|}{Beta Course ($631.5$m)} & \multicolumn{2}{c|}{CPU Usage} \\ \cline{2-11} 
    \multicolumn{1}{|c|}{} & \multicolumn{3}{c|}{APE {[}m{]}} & ME {[}m{]} & \multicolumn{3}{c|}{APE {[}m{]}} & ME {[}m{]} & \multicolumn{2}{c|}{No. of Cores} \\ \cline{2-11} 
    \multicolumn{1}{|c|}{} & max & mean & std & rmse & max & mean & std & rmse & max & mean \\ \hline
    BLAM\cite{nelson} & 3.44 & 1.01 & 0.94 & 0.43 & 3.89 & 2.27 & 0.89 & 1.27 & 1.14 & 0.93 \\ \hline
    Cartographer\cite{hess2016real} & 5.84 & 2.91 & 1.60 & 1.05 & 2.64 & 1.37 & 0.67 & 0.31 & 1.75 & 0.88 \\ \hline
    LIO-Mapping\cite{ye2019tightly} & 2.12 & 0.99 & 0.51 & 0.45 & 1.60 & 1.18 & 0.22 & 0.61 & 1.80 & 1.53 \\ \hline
    LOAM\cite{zhang2014loam} & 4.33 & 1.38 & 1.19 & 0.60 & 2.58 & 2.11 & 0.44 & 0.99 & 1.65 & 1.41 \\ \hline
    LOCUS\cite{palieri2020locus} & 0.63 & 0.26 & 0.18 & 0.28 & 1.20 & 0.58 & 0.39 & 0.48 & 3.39 & 2.72 \\ \hline
    DLO & \textbf{0.40} & \textbf{0.18} & \textbf{0.06} & \textbf{0.19} & \textbf{0.50} & \textbf{0.16} & \textbf{0.09} & \textbf{0.19} & \textbf{0.92} & \textbf{0.62} \\ \hline
    \end{tabular}
    \vspace{-5mm}
\end{table*}

\subsection{Component Evaluation}
To investigate the impact of our system's components, including keyframe-based submapping, submap normal approximation, and the reuse of data structures, we compare each component with its counterpart using the Alpha Course dataset from the Urban circuit of the DARPA Subterranean Challenge. This dataset contains LiDAR scans from a Velodyne VLP-16 sensor, in addition to IMU measurements from a VectorNav VN-100, collected across 60 minutes in an abandoned powerplant located in Elma, WA which contains multiple perceptual challenges such as large or self-similar scenes (Fig.~\ref{fig:alpha}). For these component-wise evaluations, data was processed using a 4-core Intel i7 1.30GHz CPU.

\subsubsection{Keyframe-Based Submapping}

We compared the absolute pose error (APE), processing time, and CPU load across three submapping schemes, including: radius-based ($r$ = $10$m), keyframe-based with a $1$m static threshold, and keyframe-based with adaptive thresholding. For keyframe-based variants, we used 10 nearest-neighbor and 10 convex hull keyframes for submap derivation. From Fig.~\ref{fig:ape}, the influence of our approach is clear: submapping in keyframe-space can significantly reduce positional error by considering more distant points that would otherwise be outside the scope of a radius-based approach. These additional points influence the outcome of the GICP optimization process as they are considered during error minimization for the optimal transform; this is especially important in purely frontend-based odometry, since any additional error in pose can quickly propagate over time due to drift. Processing time and CPU load showed similar trends: radius-based processed each scan notably slower at $74.2$ms per scan with an average of $37.5\%$ CPU load as compared to $21.6$ms / $10.2\%$ and $19.1$ms / $9.1\%$ for static and adaptive schemes, respectively.

\begin{figure}[!t]
    \centering
    \includegraphics[width=0.9\columnwidth]{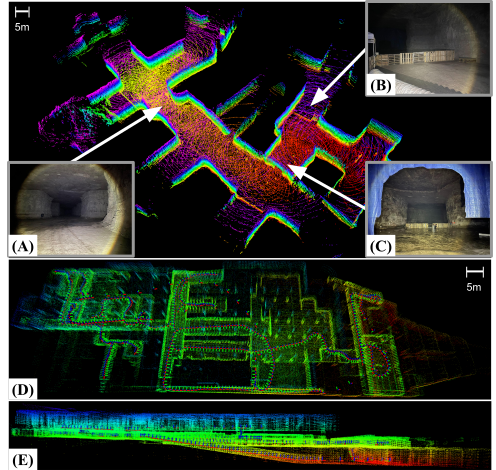}
    \vspace{-2mm}
    \caption{\textbf{Extreme environments.} \textit{Top:} A section of an underground mine in Lexington, KY mapped autonomously using our custom drone while running DLO. This environment contained challenging conditions such as: (A) low illuminance, (B) object obstructions, and (C) wet and muddy terrain. \textit{Bottom:} Top-down (D) and side (E) views of the three levels of an abandoned subway located in Downtown Los Angeles, CA mapped via DLO using a Velodyne VLP-16 on a quadruped. In this run, we manually tele-operated the legged robot to walk up, down, and around each floor for a total of $856$m.}
    \label{fig:extreme}
    \vspace{-2mm}
\end{figure}

\subsubsection{Data Structure Recycling}
To evaluate the effectiveness of data reusage, we measured and compared the processing time and CPU usage between different recycling schemes via a box plot (Fig.~\ref{fig:data_recycling}) and percentage of dropped scans over the dataset (Table~\ref{table:dropped_scans}). In a naive system which explicitly calculates each kdtree and cloud covariance, computation time exceeded LiDAR rate (10Hz for Velodyne) with a high average of $69.8$ms per scan and nearly $10\%$ of scans dropped due to high processing time. Recycling kdtrees but not covariances provides a slight improvement in processing time and CPU percentage, while recycling covariances but not kdtrees provides a more prominent performance boost; this is reasonable since our covariance recycling scheme is more aggressive than kdtree reusage. Finally, using the full scheme as detailed in Table~\ref{table:structs} significantly decreases both metrics, with an average processing time of $21.9$ms and $9.5\%$ CPU load, which prevents any LiDAR frames from dropping.

\subsubsection{NanoGICP}
To compare NanoGICP with the state-of-the-art, we use FastGICP's \cite{koide2021voxelized} benchmark alignment code found in the authors' open-source repository. This benchmark measures the average convergence time to align two LiDAR scans across $100$ runs, and we compare against PCL's \cite{Rusu_ICRA2011_PCL} GICP implementation as well as FastGICP's multithreaded implementation. Note that we do not compare against the voxelized FastGICP variant, since this method approximates planes with groups of planes and decreases overall accuracy. All tested algorithms were initialized with an identity prior, and as shown in Fig.~\ref{fig:gicp}, we observed that NanoGICP converged faster on average ($42.53$ms) when compared to FastGICP ($72.88$ms) and PCL's GICP ($178.24$ms).

\subsection{Benchmark Results}
The odometry accuracy and CPU load of DLO was compared to several LiDAR and LiDAR-IMU odometry methods --- including BLAM\cite{nelson}, Cartographer\cite{hess2016real}, LIO-Mapping\cite{ye2019tightly}, LOAM\cite{zhang2014loam}, and LOCUS\cite{palieri2020locus} --- using the Alpha and Beta course dataset from the Urban Circuit of the Subterranean Challenge  (numbers and ground truth retrieved from \cite{palieri2020locus}). We note that LIO-SAM \cite{shan2020lio} and LVI-SAM \cite{lvisam2021shan}, two state-of-the-art tightly-coupled approach, could not be tested at the time of this work due to their sensitive calibration procedure and strict input data requirements. We observed that our method's CPU load was measured to be far lower than any other algorithm, using less than one core both on average and at its peak. This is likely a result how our system derives its submap, in addition to the extensive reuse of internal data structures. This observation can also explain DLO's much lower absolute pose error (APE) and mean error (ME), with similar trends in the relative pose error. With this faster processing time, our method outperformed all other methods in both Alpha and Beta courses, having more than twice the accuracy in the Beta course for max, mean and standard deviation, even without motion distortion correction. In addition to our more permissive submapping approach, we are less likely to drop frames than other methods and have the processing capital to match the dense point clouds at a higher resolution.

\begin{figure}[!t]
    \centering
    \vspace{2mm}
    \includegraphics[width=0.85\columnwidth]{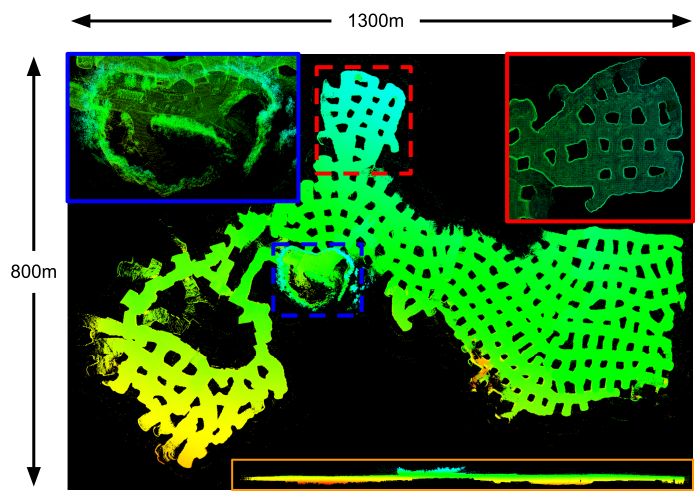}
    \vspace{-2mm}
    \caption{\textbf{Mega Cavern.} Different views of the Mega Cavern in Louisville, KY mapped by our DLO algorithm, with a total estimated trajectory of $9057.66$m. Data is courtesy of Team Explorer.}
    \label{fig:mega_cavern}
    \vspace{-2mm}
\end{figure}

\subsection{Field Experiments}
We additionally tested and implemented our solution on several custom robotic platforms for real-world field operation. Specifically, we integrated DLO onto an aerial vehicle (Fig.~\ref{fig:header}A) with an Ouster OS1 and a Boston Dynamics Spot (Fig.~\ref{fig:header}B) with a Velodyne VLP-16. Both systems contained a VectorNav VN-100 IMU rigidly mounted below the base of the LiDAR and processed data on an Intel NUC Board NUC7i7DNBE 1.9GHz CPU. We conducted both manual and autonomous traversals in two perceptually-challenging environments: in an underground limestone cave in Lexington, KY and at an abandoned subway in Los Angeles, CA (Fig.~\ref{fig:extreme}). Both locations contained environmental properties which often challenge perceptual systems, including poor lighting conditions, featureless corridors, and the presence of particulates such as dust or fog. Despite traversing over $850$m across three different levels in the abandoned subway, our system reported only a $10$cm end-to-end drift, largely owing to DLO's robust keyframing scheme which adapted to large and small spaces. Our tests in the underground mine showed similar promise: while this environment lacked any external lighting deep within the cave, DLO could still reliably track our aerial vehicle across $348$m of autonomous flight. These results demonstrate the real-world reliability of our method.


\section{Conclusion}
This work presented Direct LiDAR Odometry (DLO), a lightweight and accurate frontend localization solution with minimal computational overhead for long-term traversals in extreme environments. A key innovation which distinguishes our work from others is how we efficiently derive a local submap for global pose refinement using a database of keyframe-point cloud pairs. This in turn permits a substantial number of solver data structures to be shared and reused between system modules, all of which is facilitated using our custom NanoGICP cloud registration package. We demonstrate the reliability of our approach through benchmarks and extensive field experiments on multiple platforms operating in large-scale perceptually-challenging environments, and we invite others to use and evaluate our open-source code. DLO was developed for and used on NASA JPL's Team CoSTAR's fleet of quadrotors in the DARPA Subterranean Challenge (Fig.~\ref{fig:mega_cavern}), and in the future we are interested in tighter IMU integration as well as motion distortion correction.


\noindent \textbf{Acknowledgements:} The authors would like to thank Team CoSTAR teammates and colleagues, including Amanda Bouman, Luca Carlone, Micah Corah, Kamak Ebadi, Seyed Fakoorian, David Fan, Sung Kim, Benjamin Morrell, Joshua Ott, Andrzej Reinke, Toni Rosinol, and Patrick Spieler, for their valuable insight and productive discussions.


\bibliographystyle{IEEEtran}
\bibliography{references}

\begin{thebibliography}{10}
\providecommand{\url}[1]{#1}
\csname url@samestyle\endcsname
\providecommand{\newblock}{\relax}
\providecommand{\bibinfo}[2]{#2}
\providecommand{\BIBentrySTDinterwordspacing}{\spaceskip=0pt\relax}
\providecommand{\BIBentryALTinterwordstretchfactor}{4}
\providecommand{\BIBentryALTinterwordspacing}{\spaceskip=\fontdimen2\font plus
\BIBentryALTinterwordstretchfactor\fontdimen3\font minus
  \fontdimen4\font\relax}
\providecommand{\BIBforeignlanguage}[2]{{%
\expandafter\ifx\csname l@#1\endcsname\relax
\typeout{** WARNING: IEEEtran.bst: No hyphenation pattern has been}%
\typeout{** loaded for the language `#1'. Using the pattern for}%
\typeout{** the default language instead.}%
\else
\language=\csname l@#1\endcsname
\fi
#2}}
\providecommand{\BIBdecl}{\relax}
\BIBdecl

\bibitem{cadena2016past}
C.~Cadena, L.~Carlone \emph{et~al.}, ``Past, present, and future of
  simultaneous localization and mapping: Toward the robust-perception age,''
  \emph{IEEE Transactions on Robotics}, 2016.

\bibitem{shan2018lego}
T.~Shan and B.~Englot, ``Lego-loam: Lightweight and ground-optimized lidar
  odometry and mapping on variable terrain,'' in \emph{International Conference
  on Intelligent Robots and Systems}, 2018.

\bibitem{ebadi2020lamp}
K.~Ebadi, Y.~Chang \emph{et~al.}, ``Lamp: Large-scale autonomous mapping and
  positioning for exploration of perceptually-degraded subterranean
  environments,'' in \emph{IEEE International Conference on Robotics and
  Automation}, 2020.

\bibitem{lvisam2021shan}
T.~Shan, B.~Englot \emph{et~al.}, ``Lvi-sam: Tightly-coupled
  lidar-visual-inertial odometry via smoothing and mapping,'' in \emph{IEEE
  International Conference on Robotics and Automation}, 2021.

\bibitem{ye2019tightly}
H.~Ye, Y.~Chen, and M.~Liu, ``Tightly coupled 3d lidar inertial odometry and
  mapping,'' in \emph{International Conference on Robotics and Automation},
  2019.

\bibitem{shan2020lio}
T.~Shan, B.~Englot \emph{et~al.}, ``Lio-sam: Tightly-coupled lidar inertial
  odometry via smoothing and mapping,'' in \emph{IEEE/RSJ International
  Conference on Intelligent Robots and Systems}, 2020.

\bibitem{bhatia2010survey}
N.~Bhatia, ``Survey of nearest neighbor techniques,'' \emph{International
  Journal of Computer Science and Information Security}, 2010.

\bibitem{chen1992object}
Y.~Chen and G.~Medioni, ``Object modelling by registration of multiple range
  images,'' \emph{Image and Vision Computing}, 1992.

\bibitem{segal2009generalized}
A.~Segal, D.~Haehnel, and S.~Thrun, ``Generalized-icp.'' in \emph{Robotics:
  Science and Systems (RSS)}, 2009.

\bibitem{zhang2014loam}
J.~Zhang and S.~Singh, ``Loam: Lidar odometry and mapping in real-time.'' in
  \emph{Robotics: Science and Systems}, 2014.

\bibitem{yew20183dfeat}
Z.~J. Yew and G.~H. Lee, ``3dfeat-net: Weakly supervised local 3d features for
  point cloud registration,'' in \emph{Proceedings of the European Conference
  on Computer Vision}, 2018.

\bibitem{nelson}
\BIBentryALTinterwordspacing
E.~Nelson, ``B(erkeley) l(ocalization) a(nd) m(apping).'' [Online]. Available:
  \url{https://github.com/erik-nelson/blam}
\BIBentrySTDinterwordspacing

\bibitem{palieri2020locus}
M.~Palieri, B.~Morrell \emph{et~al.}, ``Locus: A multi-sensor lidar-centric
  solution for high-precision odometry and 3d mapping in real-time,''
  \emph{IEEE Robotics and Automation Letters}, 2020.

\bibitem{fastlio1}
W.~Xu and F.~Zhang, ``Fast-lio: A fast, robust lidar-inertial odometry package
  by tightly-coupled iterated kalman filter,'' \emph{IEEE Robotics and
  Automation Letters}, 2021.

\bibitem{fastlio2}
W.~Xu, Y.~Cai \emph{et~al.}, ``Fast-lio2: Fast direct lidar-inertial
  odometry,'' \emph{arXiv preprint arXiv:2107.06829}, 2021.

\bibitem{agha2021nebula}
A.~Agha, K.~Otsu \emph{et~al.}, ``Nebula: Quest for robotic autonomy in
  challenging environments; team costar at the darpa subterranean challenge,''
  \emph{Journal of Field Robotics}, 2021.

\bibitem{koide2021voxelized}
K.~Koide, M.~Yokozuka \emph{et~al.}, ``Voxelized gicp for fast and accurate 3d
  point cloud registration,'' in \emph{IEEE International Conference on
  Robotics and Automation}, 2021.

\bibitem{blanco2014nanoflann}
J.~L. Blanco and P.~K. Rai, ``nanoflann: a {C}++ header-only fork of {FLANN}, a
  library for nearest neighbor ({NN}) with kd-trees,''
  \url{https://github.com/jlblancoc/nanoflann}, 2014.

\bibitem{hess2016real}
W.~Hess, D.~Kohler \emph{et~al.}, ``Real-time loop closure in 2d lidar slam,''
  in \emph{IEEE International Conference on Robotics and Automation}, 2016.

\bibitem{Rusu_ICRA2011_PCL}
R.~B. Rusu and S.~Cousins, ``{3D is here: Point Cloud Library (PCL)},'' in
  \emph{{IEEE International Conference on Robotics and Automation}}, 2011.

\end{thebibliography}


\end{document}